\documentclass[letterpaper]{article} 
\usepackage{aaai2026}  
\usepackage{times}  
\usepackage{helvet}  
\usepackage{courier}  
\usepackage[hyphens]{url}  
\usepackage{graphicx} 
\usepackage{natbib}  
\usepackage{caption} 
\usepackage{algorithm}

\usepackage{amsmath,amsthm,amssymb,amsfonts}
\usepackage{bm}
\usepackage{booktabs}
\usepackage{subcaption}
\usepackage{cite}
\usepackage{color}
\usepackage{enumerate}
\usepackage{utfsym}
\usepackage{multirow}
\usepackage{makecell}
\usepackage{threeparttable}
\usepackage{array}
\newtheorem{theorem}{Theorem}
\usepackage{algpseudocode}
\usepackage{diagbox}
\usepackage{tabularx}

\usepackage{newfloat}
\usepackage{listings}
\DeclareCaptionStyle{ruled}{labelfont=normalfont,labelsep=colon,strut=off} 
\floatstyle{ruled}
\newfloat{listing}{tb}{lst}{}
\floatname{listing}{Listing}
\title{DS-ATGO: Dual-Stage Synergistic Learning via Forward Adaptive Threshold and \\ Backward Gradient Optimization for Spiking Neural Networks}

\author {
Jiaqiang Jiang\textsuperscript{\rm 1, 2}, 
Wenfeng Xu\textsuperscript{\rm 1, 2}, 
Jing Fan\textsuperscript{\rm 1, 2}, 
Rui Yan\textsuperscript{\rm 1, 2}\thanks{Corresponding author: Rui Yan (ryan@zjut.edu.cn)}
}
\affiliations {
\textsuperscript{\rm 1}College of Computer Science and Technology, Zhejiang University of Technology, Hangzhou 310023 \\
\textsuperscript{\rm 2}Zhejiang Key Laboratory of Visual Information Intelligent Processing, Hangzhou 310023 \\
\{jqjiang, wfxu, fanjing, ryan\}@zjut.edu.cn
}

\usepackage{bibentry}

\begin{document}

\maketitle

\begin{abstract}
	Brain-inspired spiking neural networks (SNNs) are recognized as a promising avenue for achieving efficient, low-energy neuromorphic computing. Direct training of SNNs typically relies on surrogate gradient (SG) learning to estimate derivatives of non-differentiable spiking activity. However, during training, the distribution of neuronal membrane potentials varies across timesteps and progressively deviates toward both sides of the firing threshold. When the firing threshold and SG remain fixed, this may lead to imbalanced spike firing and diminished gradient signals, preventing SNNs from performing well. To address these issues, we propose a novel dual-stage synergistic learning algorithm that achieves forward adaptive thresholding and backward dynamic SG. In forward propagation, we adaptively adjust thresholds based on the distribution of membrane potential dynamics (MPD) at each timestep, which enriches neuronal diversity and effectively balances firing rates across timesteps and layers. In backward propagation, drawing from the underlying association between MPD, threshold, and SG, we dynamically optimize SG to enhance gradient estimation through spatio-temporal alignment, effectively mitigating gradient information loss. Experimental results demonstrate that our method achieves significant performance improvements. Moreover, it allows neurons to fire stable proportions of spikes at each timestep and increases the proportion of neurons that obtain gradients in deeper layers.
\end{abstract}
\begin{links}
	\link{Code}{https://github.com/jqjiang1999/DS-ATGO}
\end{links}

\section{Introduction}
As a new paradigm that combines biological plausibility with computational efficiency, spiking neural networks (SNNs) have recently attracted widespread attention \cite{maass1997networks}. Unlike artificial neural networks (ANNs), which work with continuous activation and real-valued encoding \cite{ju2025m2llm,ju2025uni,zhang2025dynamic,feng2025bimark,zhang2024unraveling}, SNNs operate with asynchronous, discrete spiking signals.
\begin{figure}[htbp]
	\centering
	\includegraphics[width=0.95\linewidth]{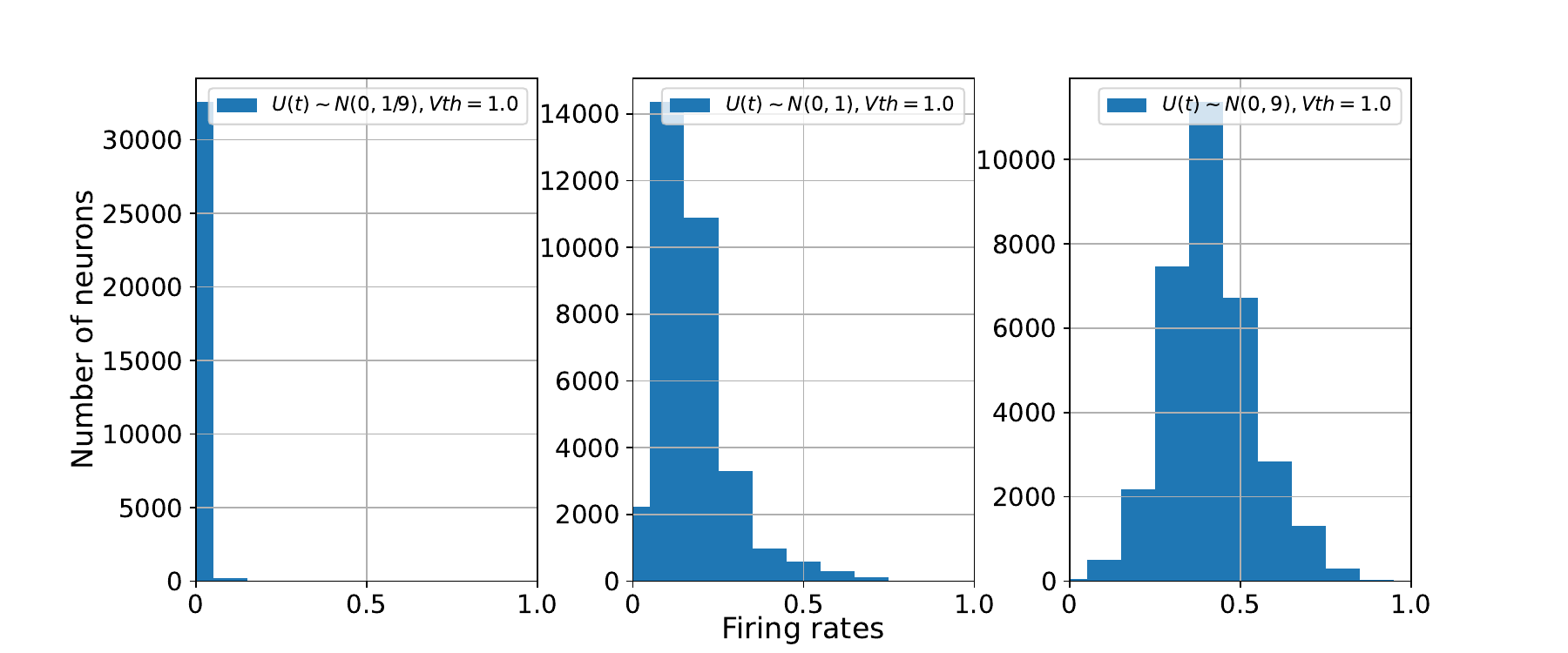}
	\caption{The distributions of firing rates at differ variances of membrane potential when $V_{th}=1.0$ \cite{zheng2021going}.}
	\label{Fig::Vth_differentVar}
\end{figure}
This spike train-based way renders SNNs highly promising for spatio-temporal (ST) information processing and energy-efficient computation compared to ANNs \cite{roy2019towards}. Despite these advantages, the non-differentiable nature of spike activity hinders gradient backpropagation, making it more difficult to directly train high-performance SNNs than ANNs. To overcome this issue, surrogate gradient (SG) learning \cite{wu2018spatio} estimates the gradients of output signals by introducing continuous smooth functions, which reconstruct a complete propagation path of ST gradients.

\begin{figure}[htbp]
	\centering
	\includegraphics[width=0.93\linewidth]
	{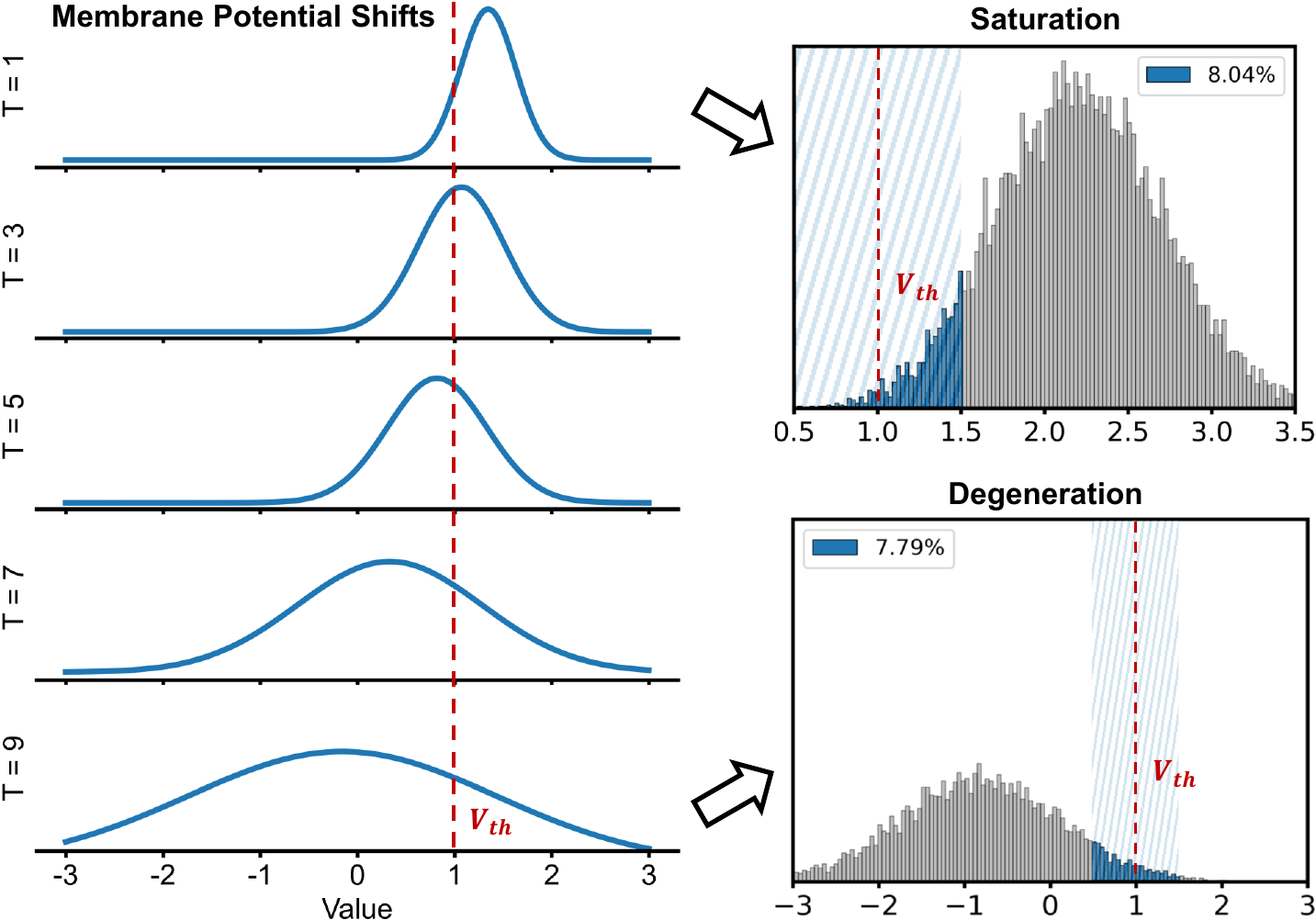}
	\caption{The distribution of membrane potentials deviating from the threshold in a vanilla SNN with ten timesteps. When almost all the membrane potentials of neurons are beyond  $V_{th}$, called saturation. Conversely, called degeneration.}
	\label{Fig::gradient_vanishing_problem}
\end{figure}

However, as spikes propagate through layers, the distribution of membrane potentials shifts and may fall into inappropriate areas \cite{guo2022recdis,liu2025deeptage} (Fig.~\ref{Fig::gradient_vanishing_problem}). To our best knowledge, most existing SG-based methods for directly training SNNs use a fixed threshold and SG. Thus, membrane potential dynamics (MPD) may present two challenges that lead to training difficulties.

The first challenge is the imbalanced spike firing of SNNs with a fixed threshold. The neuron fires a spike only when its membrane potential exceeds the threshold $V_{th}$. In Fig.~\ref{Fig::Vth_differentVar}, when membrane potentials are too small compared to the threshold, excessive sparse firing in neurons will lead to the \textit{``spike vanishing problem"}. When membrane potentials are too large, neurons will be over-firing, reducing the ability to represent the input pattern differentially. To maintain the stability of information flow in SNN, we need to balance the threshold and membrane potentials to keep neurons in a moderately active state.
Neuroscience has observed in various brain regions that the thresholds of biological neurons are not constant but exhibit variability between and within neurons \cite{azouz2000dynamic,farries2010dynamic}. This phenomenon, known as threshold plasticity, can be regarded as an adaptation to membrane potentials, which plays an essential role in maintaining neuronal firing homeostasis \cite{fontaine2014spike}. To model threshold plasticity, \cite{wang2022LTMD,rathi2023diet,sun2024synapse,hasssan2024spiking} directly set the threshold as a learnable parameter and optimize it dynamically via gradients. Although this effectively enhances neuronal firing levels, its optimization relies solely on the gradient of thresholds itself, leaving uncertainty in maintaining the balance of firing rates in SNNs.

The other challenge is the diminished gradient information of SNNs with a fixed SG. The neuron obtains a gradient only when its membrane potential falls within the gradient-available interval of SG. In Fig.~\ref{Fig::gradient_vanishing_problem}, when membrane potentials is too small or too large compared to the threshold, the limited gradient-available interval of fixed SG causes the membrane potentials of many neurons to fall into the areas where the approximate derivatives are zero or a tiny value. In this case, only a few neurons contribute gradients, leading to the \textit{``gradient vanishing problem"}. \cite{lian2023learnable} dynamically adjusted the SG based on changes in the MPD distribution. It ignores the dynamic nature of evolving MPD in the temporal dimension, which may exacerbate the inaccurate estimation of gradients. Then,  \cite{jiang2025adaptive,liu2025deeptage} optimized the SG of different timesteps in a temporal-aligned manner. In SG learning, the gradient-available interval of SG is symmetrically centered around the threshold. However, these methods overlook the association between MPD, threshold, and SG, making SG fail to accurately capture the dynamic deviations of membrane potentials relative to thresholds. Therefore, the neuronal threshold can be regarded as an idealized tunable parameter that regulates spike firing and affects gradient optimization.

\begin{figure}[!ht]
	\centering
	\includegraphics[width=0.93\linewidth]{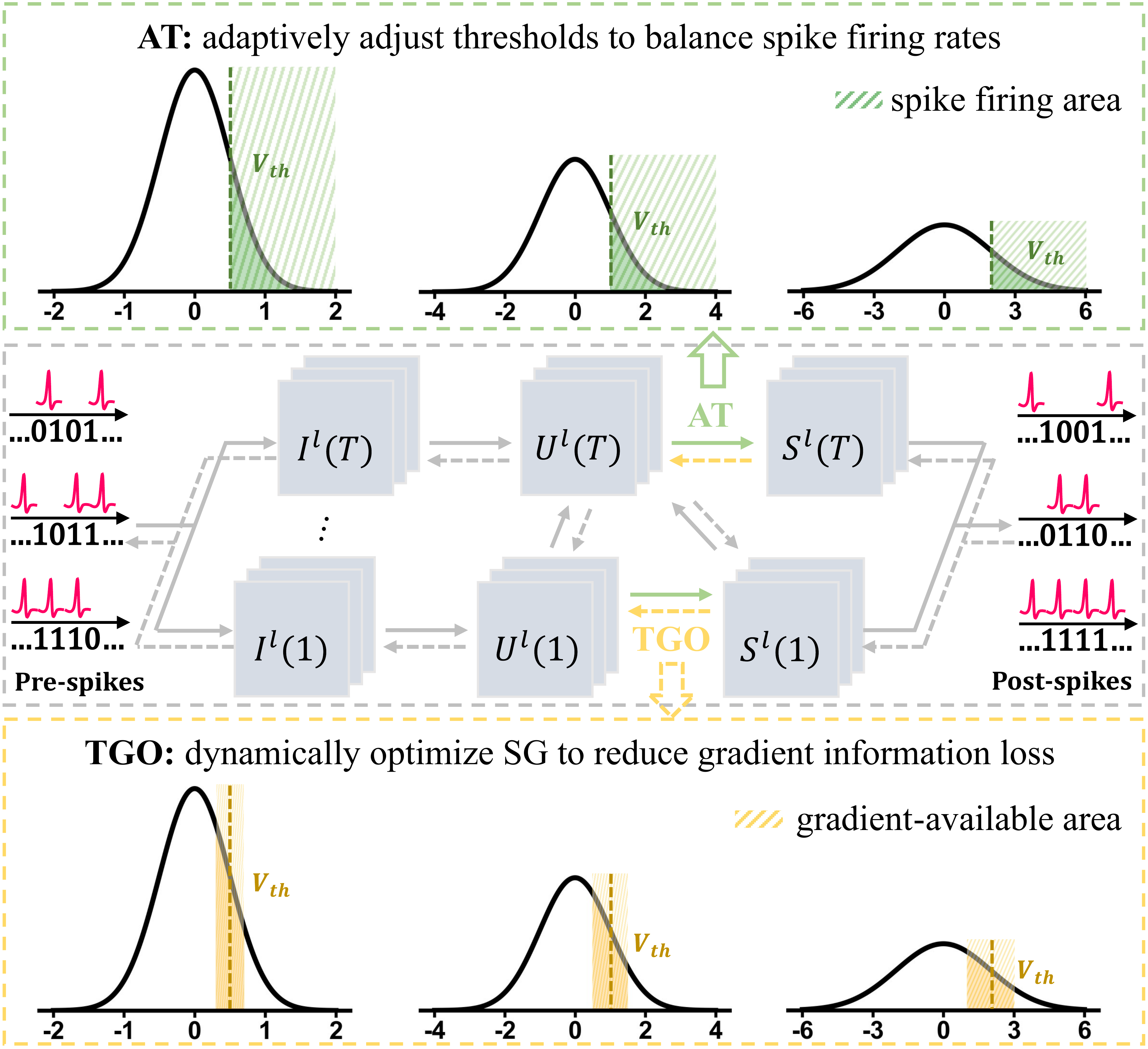}
	\caption{The overall framework of DS-ATGO. Internal dynamics of LIF neurons in a layer (gray). In forward propagation, the adaptive threshold (AT) mechanism promotes neurons to generate stable firing rates under different MPD distributions (green). In backward propagation, the threshold-driven SG optimization (TGO) method dynamically scales SG to respond to evolving MPD (yellow).}
	\label{Fig::overall_framework}
\end{figure}

In this paper, we propose a novel dual-stage synergistic learning via forward adaptive threshold and backward gradient optimization for SNNs, named DS-ATGO. By utilizing the distribution characteristics of MPD at each timestep, we adaptively adjust thresholds to enhance the ability of neurons to encode dynamic information, achieving a linear response to input signals. Building on the observation that adaptive thresholds reflect shifts in membrane potentials, we further establish a correlation between SG and MPD via thresholds. Based on that, we dynamically adjust SG in a threshold-driven manner to align with evolving MPD, contributing smoother spatio-temporal gradients and maintaining the optimal gradient domain across timesteps. The overall framework of DS-ATGO is illustrated in Fig.~\ref{Fig::overall_framework}. The main contributions of our work are summarized as follows:
\begin{itemize}
	\item We propose an adaptive threshold mechanism that adapts the firing threshold in the temporal dimension by leveraging the MPD distribution. It balances neuronal spike firing across layers, enhancing the cross-layer information transmission capability of SNNs.
	\item We propose a threshold-driven gradient optimization method that dynamically adjusts SG at each timestep by associating SG with evolving MPD via adaptive thresholds, effectively enhancing ST gradient information.
	\item The experimental results on both static and neuromorphic datasets demonstrate that DS-ATGO achieves outstanding performance with low latency without additional inference overhead.
\end{itemize}

\section{Related Works}
\subsection{Spiking Threshold Plasticity}
Spiking threshold plasticity is a biological mechanism by which neurons dynamically adjust their firing thresholds according to historical activity patterns. To express this characteristic, \cite{ai2025cross} dynamically adjusted the threshold based on the repetitive or high-frequency discharge responses of neurons, effectively preventing excessive activation of neurons. \cite{fu2025adaptation} proposed a spatio-temporal threshold adjustment strategy, which enhances the spike coding capacity of SNNs by coupling with neural dynamics. In addition, some studies on SNN training have introduced learning dynamics into spike processes \cite{wang2022LTMD,rathi2023diet}. \cite{sun2024synapse} proposed a synapse-threshold synergistic learning, which allows stable signal transmission through appropriate firing rates by simultaneously training weights and thresholds. Considering the weights and threshold same landscape makes the learning sub-optimal, \cite{hasssan2024spiking} achieved individually adaptive optimization of layer-wise thresholds and weights by introducing a separate gradient path.

\subsection{Gradient Optimization}
The concept of gradient optimization in SNN aims to promote the efficient propagation of error signals in the ST domain by adjusting SG \cite{fang2021deep}. \cite{guo2022loss} approximated the spike activity gradient by a continuously differentiable evolving asymptotic function, bridging the gap between pseudo and natural derivatives. \cite{che2022differentiable} proposed a differentiable gradient search method to optimize SG locally. \cite{lian2023learnable} unlocked the width limitation of SG based on changes in MPD distributions, increasing the proportion of neurons that obtained gradients. \cite{wang2023adaptive} adaptively learned the accurate gradients of the loss landscape in SNN by fusing the learnable relaxation degree into a prototype network with random spike noise, effectively eliminating the smoothing error of SG learning. \cite{wang2025potential} proposed a parametric SG strategy to control and calibrate SG, which mitigates the degradation caused by improper selection of SG. \cite{jiang2025adaptive,liu2025deeptage} promotes balanced optimization by enhancing the gradients in a temporal-aligned manner.

\paragraph{Motivation:} Overall, although existing methods achieve threshold or SG adjustment, they ignore the intrinsic association among membrane potential, threshold, and SG, which have limitations in jointly solving the two problems caused by membrane potential shifts. This motivates us to achieve the synergistic optimization of thresholds and SG to maintain balanced firing rates and stable gradient flows in SNNs.

\section{Preliminaries}
\subsection{Spiking Neural Model}
In this work, we use the Leaky Integrate-and-Fire (LIF) model in SNNs. The dynamics of an iterative LIF neuron \cite{wu2019direct} can be modeled by
\begin{align}
	I_i^l(t) &= \sum_{j=1}^{N(l-1)} w_{ij}^l S_j^{l-1}(t), \\
	U_i^l(t) &= \tau U_i^l(t-1) \odot (1 - S_i^l(t-1))+ I_i^l(t), \label{Eq::iterative_LIF} \\
	S_i^l(t) &= \Theta(U_i^l(t)) = \begin{cases} 1, & U_i^l(t) \ge V_{th} \\ 0, & otherwise \end{cases} \label{Eq::activation_function}
\end{align}
where the upper $l$, subscripts $i$ and $t$ denote the $l$-th layer, $i$-th neuron, and $t$-th timestep, respectively. $N(l-1)$ denotes the number of neurons in the $(l-1)$-th layer. $w_{ij}^l$ denotes the synapse weight from the $j$-th neuron in the $(l-1)$-th layer to the $i$-th neuron in the $l$-th layer. $I$, $U$, and $S$ denote the pre-synaptic input, the membrane potential, and the output spike of neurons. $V_{th}$ is the firing threshold. $\tau$ is the membrane time constant that indicates how quickly the potential decays over time, affecting the duration of neuronal excitation.

\subsection{Surrogate Gradient Function}
In Eq. \ref{Eq::activation_function}, the firing function $\Theta(\cdot)$ of SNNs is a Heaviside function. The derivative of output signals $\frac{\partial S}{\partial U}$ is a Dirac function that tends to infinity at the threshold $V_{th}$ and zeros otherwise. Here, we employ the rectangular function \cite{wu2018spatio} to estimate the gradients of $\frac{\partial S}{\partial U}$, which is defined as
\begin{equation}
	\frac{\partial S^l_i(t)}{\partial U^l_i(t)} \thickapprox h(U^l_i(t)) = \frac{1}{\kappa}sign(|U^n_i(t)-V_{th}| \le \frac{\kappa}{2}),
	\label{SG_function}
\end{equation}
where hyperparameter $\kappa$ controls the width of $h(\cdot)$ to ensure that it integrates to 1, normally set to 1. The gradient $\frac{1}{\kappa}$ is available when the membrane potential $U^n_i(t)$ falls within the interval $[V_{th} - \frac{\kappa}{2}, V_{th} + \frac{\kappa}{2}]$.

\section{Methods}
\subsection{Adaptive Threshold Mechanism}
It is difficult for SNNs with a fixed threshold to adapt to variations in input strength, making the network prone to silence or excessive firing when processing ST dynamic data. \cite{zheng2021going} proposed a threshold-dependent batch normalization that normalized the pre-synaptic input $I(t)$ to $N(0, V_{th}^2)$ instead of $N(0, 1)$. Although this method balances pre-synaptic inputs and thresholds to increase neuronal spike activity, the intrinsic dynamics of neurons still hinder it from maintaining the stability of firing rates across the entire timestep. That motivates us to explore how to adaptively adjust the firing threshold to keep neurons homeostatic, enhancing the dynamic responsiveness of SNNs.

Recent studies have revealed that the threshold of biological neurons exhibits a positive correlation with membrane potentials \cite{pena2002postsynaptic,azouz2003adaptive,ding2022biologically}. Based on the insight that firing rates are directly regulated by membrane potential and threshold, we can adjust the threshold according to the evolving MPD, promoting neurons to fire appropriate spikes in SNNs. To achieve this, the key lies in balancing the proportion of membrane potentials that exceed the threshold. \textbf{Theorem 1} shows that when the threshold is set as the sum of expectation and standard deviation of MPD distributions ($V_{th}=\mu+\sigma$), the proportion of membrane potentials exceeding the threshold remains relatively constant under different distributions. Establishing a positive correlation between thresholds and membrane potentials retains the adaptive properties of biological neurons while ensuring a stable firing rate in SNNs.

\begin{theorem} 
	For $U(t)$ that satisfies a normal distribution $N(\mu,\sigma^2)$, the probability that a random variable $U_i(t)$ exceeds $\mu+\sigma$ is relatively constant and given by $P(U_i(t) > \mu+\sigma)=1-\varPhi(1)$, where $\varPhi(\cdot)$ denotes the cumulative distribution function of standard normal distributions.
\end{theorem}

\begin{figure}[htbp]
	\centering
	\includegraphics[width=0.90\linewidth]{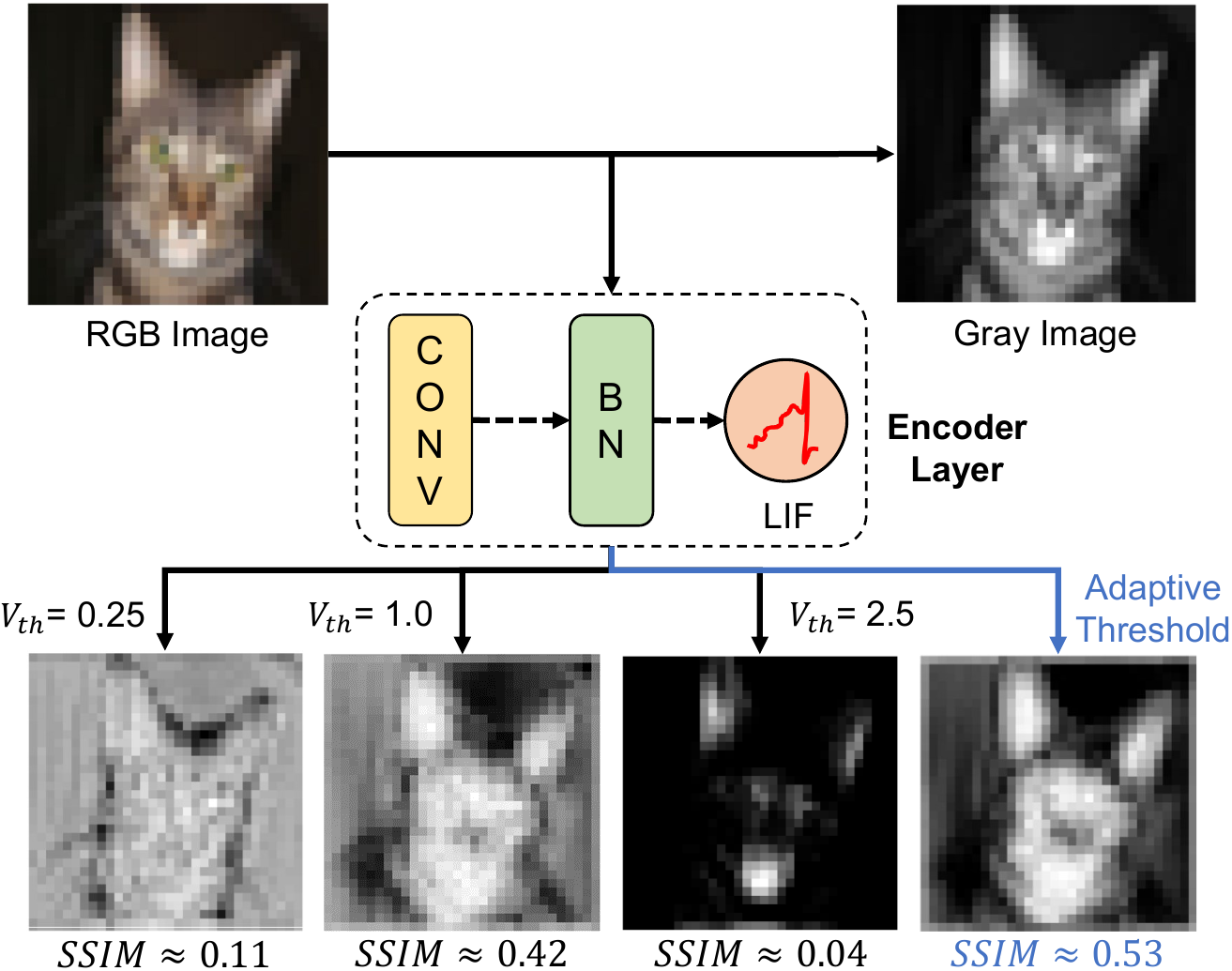}
	\caption{The structural similarity between the gray image and encoded images at different thresholds.}
	\label{Fig::dynamic_threshold}
\end{figure}

\begin{proof}
	The proof of Theorem 1 is given in \textbf{Appendix~\ref{Sec::proofs_theorems}}.
\end{proof}

To maintain neurons in a moderately active state, we first need to analyze the detailed distributions of membrane potentials. \cite{zheng2021going} derived a high degree of similarity between the distribution of pre-synaptic inputs and membrane potentials. \cite{lian2023learnable,jiang2025adaptive} extended this theorem that for a given pre-synaptic inputs $I(t) \sim N(0, V_{th}^2)$, the distribution of membrane potentials satisfies $U(t) \sim N(0, (1 + \tau^2)V_{th}^2)$. Given the approximate nature of MPD distributions derived by \cite{lian2023learnable,jiang2025adaptive}, we can also directly use the true MPD distribution during training. Moreover, to trade-off energy efficiency and performance, we introduce a factor $f_c$ to control spiking firing rates. We will discuss these two distributions and the effect of factor $f_c$ in the ablation study. Finally, the adaptive threshold mechanism can be formulated as
\begin{align}
	\Delta{V_{th}}^l(t)_n &\approx f_c * \sqrt{(1+\tau_n^l{^2}))}V_{th}, \# Estimated 	\label{Eq::adaptive_threshold}
	\\
	&= f_c *(\mathbb{E}(U^l(t)_n) +\sqrt{\mathbb{VAR}(U^l(t)_n)}), \# True \nonumber
\end{align}
where the subscript $n$ denotes the $n$-th mini-batch, $\tau$ is layer-wise learnable \cite{fang2021incorporating}. In this way, all neurons in different layers will have distinct thresholds across different timesteps. Fig.~\ref{Fig::dynamic_threshold} shows that using the proposed adaptive threshold in the encoding layer produces better quality images than the fixed threshold.

Considering that different mini-batches in the inference stage may cause fluctuations in MPD distributions that affect the threshold, inspired by BN \cite{ioffe2015batch}, we also use the moving average of thresholds during training for inference to stabilize neuron outputs. It can be described as
\begin{equation}
	\Delta{V_{th}}^l(t) = m * \Delta{V_{th}}^l(t)_n + (1-m) * \Delta{V_{th}}^l(t), 
\end{equation}
where $m$ is the momentum coefficient, which we set to 0.1.

\subsection{Threshold-driven Gradient Optimization}
\begin{figure}[htbp]
	\centering
	\includegraphics[width=1.0\linewidth]{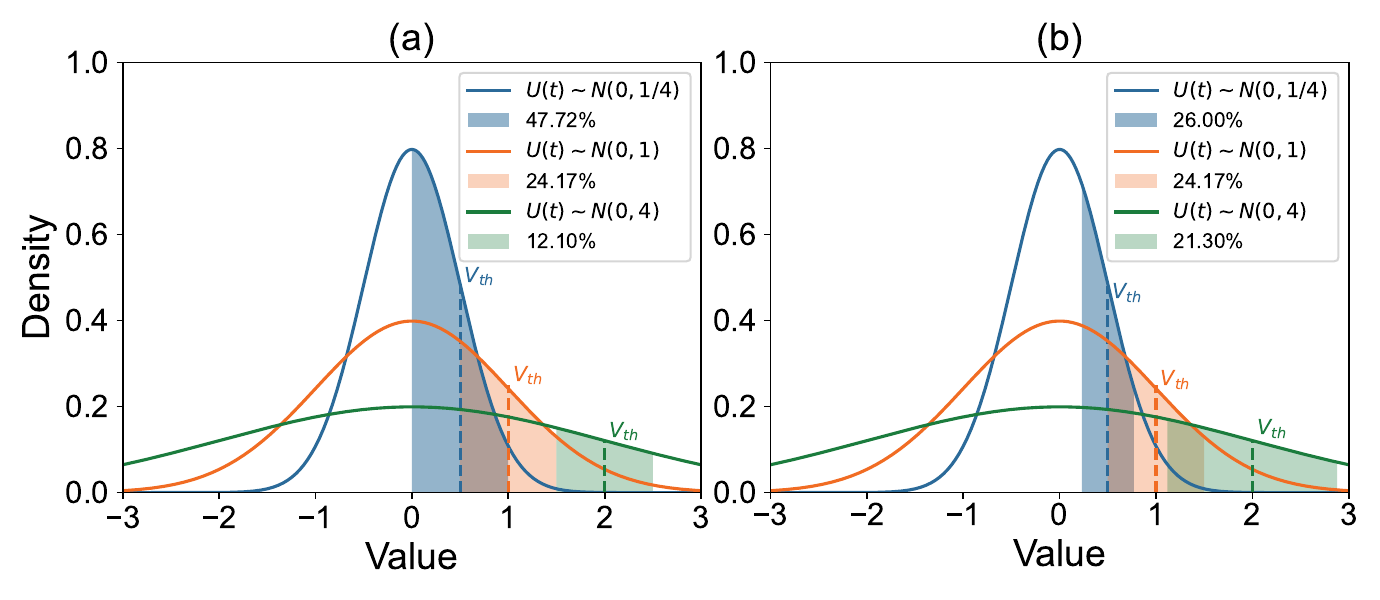}
	\caption{(a) The proportion of membrane potentials that fall into the gradient-available interval with fixed SG ($k=1$) under different distributions when using adaptive threshold. (b) The proportion of gradient-available when using the adaptive threshold and threshold-driven gradient optimization.}
	\label{Fig::gradient_match}
\end{figure}
In SNNs, adequate gradient information is crucial for achieving efficient learning, as it directly governs weight updates and model convergence. When SG remains fixed, membrane potential shifts will cause decreased gradients during backpropagation. The adaptive threshold mechanism adjusts the gradient-available interval of fixed SG to $[\Delta{V_{th}} - \frac{1}{2}, \Delta{V_{th}} + \frac{1}{2}]$ instead of $[V_{th} - \frac{1}{2}, V_{th} + \frac{1}{2}]$, but the above issue remains due to the inherent mismatch between fixed SG and evolving MPD. Specifically, in Fig.~\ref{Fig::gradient_match}, when the variance of MPD distributions becomes smaller (blue), a large number of neuronal membrane potentials will fall within the gradient-available interval for gradient computation, enlarging the approximation error with the natural gradient. Conversely, when the variance of MPD distributions becomes larger (green), only a few neurons will obtain gradients, and most neurons will fall into the saturation area with zero gradients, resulting in vanishing gradient.

To optimize SG learning, we need to modify SG accordingly to better respond to evolving MPD. \cite{lian2023learnable,jiang2025adaptive} indicated an association between SG and MPD. Given that the adaptive threshold reflects changes in MPD distributions (Eq.~\ref{Eq::adaptive_threshold}) and that the threshold determines the location of SG. Then, we propose a threshold-driven gradient optimization method that dynamically adjusts SG width (Fig.~\ref{Fig::gradient_match}), ensuring that the gradient-available interval aligns with MPD. For the above two distinct cases, we design two rules to calculate the SG width, as follows:
\begin{itemize}
	\item $\Delta{V_{th}} < V_{th}$: When the adaptive threshold $\Delta{V_{th}}$ is less than the initial threshold $V_{th}$, which indicates the MPD distribution tends to concentrate, thus we need to reduce the SG width to decrease the proportion of neurons fall into the gradient-available interval, suppressing the cumulative enlargement of gradient approximation errors.
	\item $\Delta{V_{th}} \ge V_{th}$: When the adaptive threshold $\Delta{V_{th}}$ is greater than the initial threshold $V_{th}$, which indicates the MPD distribution tends to disperse, thus we need to expand the SG width to increase the proportion of neurons falling into the gradient-available interval, mitigating the loss of gradient information.
\end{itemize}

In summary, the threshold-driven gradient optimization method can be mathematically written as
\begin{equation}
	\begin{aligned}
		k =
		\begin{cases} 
			(1 - tanh(V_{th} - \Delta{V_{th}})) * k, & \Delta{V_{th}} < V_{th} \\
			(1 + tanh(\Delta{V_{th}} - V_{th})) * k, & \Delta{V_{th}} \ge V_{th}
		\end{cases}
	\end{aligned}
\end{equation}

\subsection{The Overall Procedure of SNNs}
In the output layer, we only accumulate the membrane potential of output neurons without leakage and firing \cite{rathi2023diet}, which can be described by
\begin{equation}
	U_i^{l^o} = \frac{1}{T} \sum^T_{t=1} \sum_{j=1}^{N(l^o-1)} w^{l^o}_{ij}S^{l^o-1}_j(t), i\in\{1,2,...,c\}
\end{equation}
where $l^o$ and $c$ denote the output layer and the number of classes, respectively. As $U^l_i(t)$ not only contributes to $S^l_i(t)$ but also governs $U^l_i(t+1)$, it can be derived by
\begin{align}
	\frac{\partial L}{\partial U^l_i(t)} &= \frac{\partial L}{\partial S^l_i(t)} \frac{\partial S^l_i(t)}{\partial U^l_i(t)} + \frac{\partial L}{\partial U^l_i(t+1)} \frac{\partial U^l_i(t+1)}{\partial U^l_i(t)}, \\
	\frac{\partial L}{\partial S^l_i(t)} &= \frac{\partial L}{\partial U^l_i(t+1)} \frac{\partial U^l_i(t+1)}{\partial S^l_i(t)} \nonumber \\ 
	& + \sum^{N(l+1)}_{j=1} \frac{\partial L}{\partial U^{l+1}_j(t)} \frac{\partial U^{l+1}_j(t)}{\partial S^l_j(t)}.
\end{align}
where $\frac{\partial S^l_i(t)}{\partial U^l_i(t)}$ are approximated by Eq.~\ref{SG_function}. Then, the gradient of synaptic weights $w^l_{ij}$ can be derived by the chain rule:
\begin{align}
	\frac{\partial L}{\partial w^l_{ij}} &= \sum^T_{t=1} \frac{\partial L}{\partial U^l_i(t)} \frac{\partial U^l_i(t)}{\partial I^l_i(t)} \frac{\partial I^l_i(t)}{\partial w^l_{ij}} \nonumber \\
	&= \sum^T_{t=1} \frac{\partial L}{\partial U^l_i(t)} \sum^{N(l-1)}_{j=1}S^{l-1}_j(t).
\end{align}
Moreover, the pseudocode of the overall procedure is briefed in \textbf{Appendix~\ref{Sec::overall_procedure}}.

\section{Experiments}
\subsection{Ablation Study}
\begin{figure}[ht]
	\centering
	\includegraphics[width=0.95\linewidth,height=7cm]{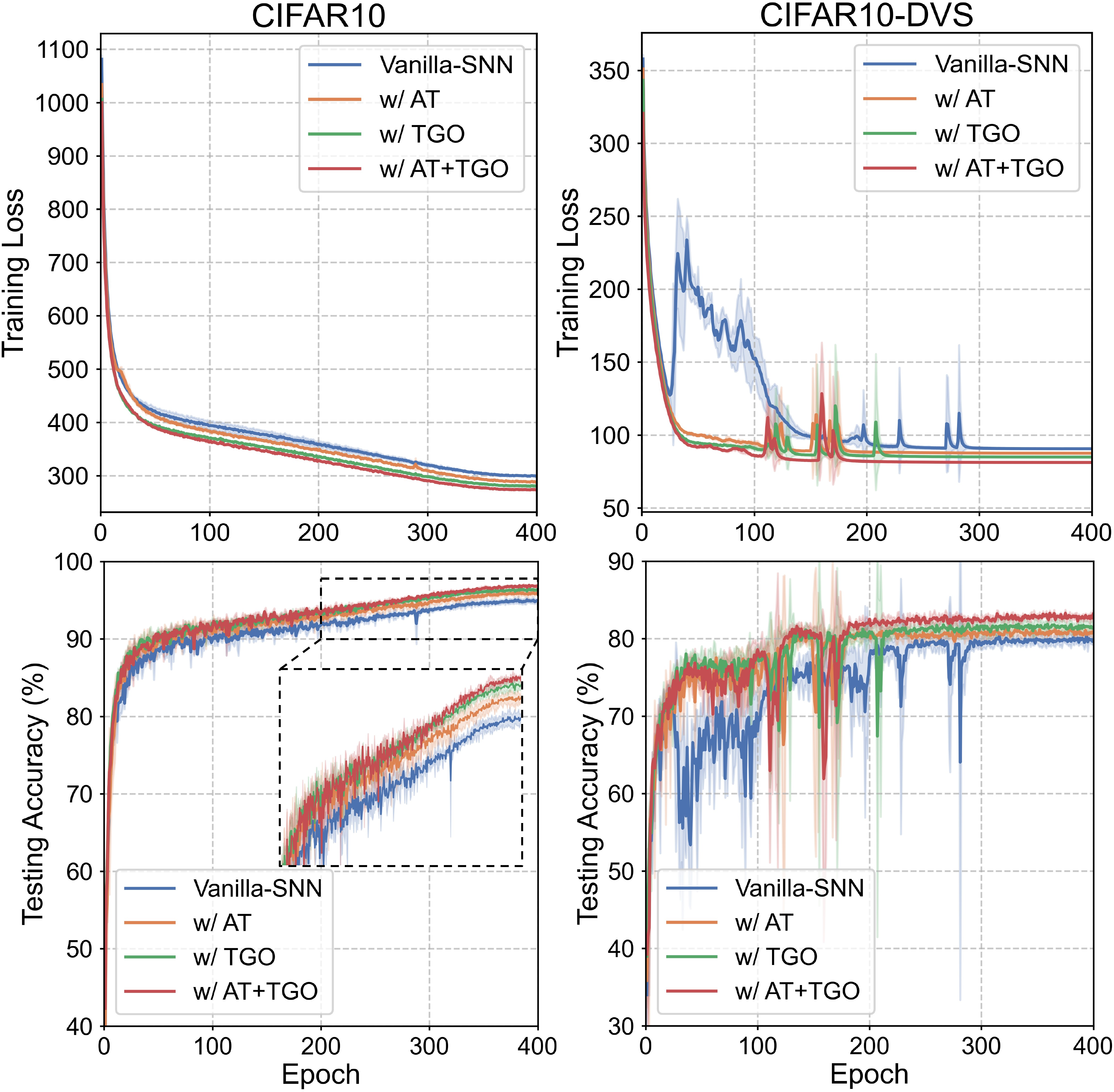}
	\caption{Comparison of training loss and test accuracy.}
	\label{Fig::accuracu_loss}
\end{figure}

The performance improvement of our model benefits from the adaptive threshold mechanism (AT) and threshold-driven gradient optimization method (TGO). We conducted experiments to evaluate their contributions. In Fig.~\ref{Fig::accuracu_loss}, applying AT (w/~AT) outperforms Vanilla-SNN on two datasets. This indicates that adaptively adjusting thresholds in response to evolving MPD enables flexible control over the timing and intensity of neuron activation, enhancing the SNN's ability to model complex dynamic information. Notably, applying TGO (w/~TGO) also achieves significant improvements over Vanilla-SNN, even surpassing AT.
\begin{figure}[ht]
	\centering
	\includegraphics[width=0.75\linewidth,height=4.3cm]{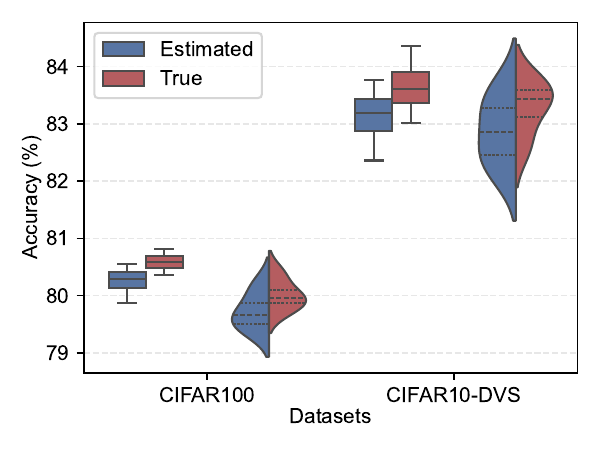}
	\caption{Comparison of estimated with true MPD distribution and w/ vs. w/o moving average (box plot vs. violin plot).}
	\label{Fig::ablation_study_ETMA}
\end{figure}
The reason is that, unlike AT, which focuses on spike modulation in information encoding, TGO synchronously adjusts SG to accurately capture the dynamic deviation of membrane potentials relative to thresholds. It effectively maintains the balance of gradient signals in the temporal dimension, enhancing the efficiency and correctness of parameter updates. Furthermore, combining the two techniques (w/~AT+TGO) enables SNNs to balance firing rates for improved encoding efficiency and optimize SG through temporal alignment for gradient enhancement, further boosting performance. It achieved remarkable improvements of 1.71\% and 2.30\% on the CIFAR10 and CIFAR10-DVS datasets, respectively.

Moreover, Fig.~\ref{Fig::ablation_study_ETMA} illustrates that DS-ATGO driven by the true MPD distribution outperforms the theoretical distribution derived by \cite{lian2023learnable}, which indicates that the theorem derivation involves certain approximation errors. Then, we evaluated the effectiveness of the moving average rule. When applying this rule in AT, the accuracy of both datasets surpassed the baseline without the rule. Especially for CIFAR10-DVS trained with mini-batches, the threshold moving average more effectively adapts to MPD fluctuations, exhibiting higher performance improvements.
\begin{table*}[!ht]
	\centering
	\setlength{\tabcolsep}{1mm}
	\renewcommand{\arraystretch}{0.7}
	\begin{tabular}{c|cccccc}
		\toprule
		\textbf{Dataset}        & \textbf{Method}        & \textbf{AT}     & \textbf{GO}  & \textbf{Architecture}    & \textbf{Timestep}     & \textbf{Accuarcy(\%)} \\
		\midrule
		\multirow{8}*{CIFAR10}
		& STAtten+ \cite{lee2025spiking}     & \usym{2717}   & \usym{2717}  & Spikformer-4-384  & 4     & 94.36 \\
		
		& Distillation-based SNN \cite{yu2025efficient}     & \usym{2717}   & \usym{2717}  & ResNet-18  & 6 / 4    & 95.96 / 95.57 \\
		
		
		& STL-SNN \cite{sun2024synapse}     & \usym{2714}   & \usym{2717}  & 8-layers SNN  & 8     & 92.42 \\
		
		& LT-SNN \cite{hasssan2024spiking}     & \usym{2714}   & \usym{2717}  & Spikformer-4-256  & 4     & 95.19 \\
		
		
		
		
		
		& DeepTAGE \cite{liu2025deeptage}          & \usym{2717}   & \usym{2714}    & ResNet-18 & 4   & 95.86 \\
		
		& MPD-AGL \cite{jiang2025adaptive}          & \usym{2717}   & \usym{2714}   & ResNet-19  & 4 / 2     & 96.35 / 96.18 \\
		
		& AT-LIF$\&$ASG-S \cite{ai2025cross}          & \usym{2714}   & \usym{2714}    & ResNet-18 & 6 / 4   & 95.47 / 95.30 \\
		
		& \textbf{Ours}          & \usym{2714}   & \textbf{\usym{2714}}     & \textbf{ResNet-19}  & \textbf{2}     & \textbf{96.91}$\pm$0.12 \\
		
		\midrule
		\multirow{8}*{CIFAR100}
		& TMC \cite{yan2025training}     & \usym{2717}   & \usym{2717}  & ResNet-19  & 6 / 4 / 2     & 78.05 / 77.52 / 76.35 \\
		
		& SNN-ViT \cite{wang2025spiking}     & \usym{2717}   & \usym{2717}  & VGG-16  & 4     & 80.01 \\
		
		
		
		& LT-SNN \cite{hasssan2024spiking}     & \usym{2714}   & \usym{2717}  & ResNet-19  & 6 / 2   & 74.82 / 72.78 \\
		
		& ALSF \cite{fu2025adaptation}     & \usym{2714}   & \usym{2717}  & ResNet-19  & 2     & 78.73 \\
		
		
		
		
		& DeepTAGE \cite{liu2025deeptage}          & \usym{2717}   & \usym{2714}    & ResNet-18 & 4   & 78.80 \\
		
		& MPD-AGL \cite{jiang2025adaptive}          & \usym{2717}   & \usym{2714}   & ResNet-19     & 4 / 2    & 79.72 / 78.84 \\
		
		& AT-LIF$\&$ASG-S \cite{ai2025cross}          & \usym{2714}   & \usym{2714}    & ResNet-18 & 6 / 4   & 76.44 / 76.42 \\
		
		& \textbf{Ours}          & \usym{2714} & \usym{2714}  & \textbf{ResNet-19}     & \textbf{2}     & \textbf{80.59}$\pm$0.17 \\
		
		\midrule
		\multirow{8}*{CIFAR10-DVS}
		
		
		& FSTA-SNN \cite{yu2025fsta} & \usym{2717} & \usym{2717}   & ResNet-20       & 10     & 81.50 \\
		
		& QP-SNN \cite{wei2025qpsnn} & \usym{2717} & \usym{2717}   & VGGSNN       & 10     & 82.10 \\
		
		& STL-SNN \cite{sun2024synapse}       & \usym{2714}   & \usym{2717}  & 7-layers SNN   & 20     & 77.30 \\
		
		& LT-SNN \cite{hasssan2024spiking}     & \usym{2714}   & \usym{2717}  & VGG-9  & 30   & 80.07 \\
		
		
		
		
		& DeepTAGE \cite{liu2025deeptage}          & \usym{2717}   & \usym{2714}    & VGG-11 & 10   & 81.23 \\
		
		& MPD-AGL \cite{jiang2025adaptive}          & \usym{2717}   & \usym{2714}   & VGGSNN  & 10   & 82.50 \\
		
		& AT-LIF$\&$ASG-S \cite{ai2025cross}          & \usym{2714}   & \usym{2714}    & VGGSNN & 10   & 78.70 \\
		
		& \textbf{Ours}  & \usym{2714}   & \usym{2714} & \textbf{VGGSNN}     & \textbf{10}     & \textbf{83.70}$\pm$0.41\\
		\midrule
		\multirow{6}*{ImageNet}
		& AGMM \cite{liang2025towards} & \usym{2717} & \usym{2717}   & ResNet-18       & 4     & 64.67 \\
		
		& ReverB \cite{guo2025reverbsnn} & \usym{2717} & \usym{2717}   & ResNet-18       & 4     & 66.22 \\
		
		& MTT \cite{du2025temporal} & \usym{2717} & \usym{2717}   & ResNet-34       & 4     & 67.54 \\
		
		
		& LTF$\&$ALSF \cite{fu2025adaptation}       & \usym{2714}   & \usym{2717}  & SEW ResNet-18   & 4     & 63.67 \\
		
		& DeepTAGE \cite{liu2025deeptage}       & \usym{2717}   & \usym{2714}  & ResNet-18   & 4     & 68.52 \\
		
		& \textbf{Ours}  & \usym{2714}   & \usym{2714}     & \textbf{ResNet-18} & \textbf{4}     & \textbf{68.86}$\pm$0.25 \\ 
		\bottomrule
	\end{tabular}
	\caption{Comparison of classification accuracy. \textbf{AT} and \textbf{GO} denote adaptive threshold and gradient optimization, respectively.}
	\label{Tab::classification_accuracy}
\end{table*}
\subsection{Performance Evaluation}
As listed in Table~\ref{Tab::classification_accuracy}, we compare the classification accuracy of our method with other advanced methods. DS-ATGO performed well on all four datasets, achieving an accuracy of 96.91\%/80.59\% with low latency on CIFAR10/100, 83.70\% and 68.86\% accuracy on CIFAR10-DVS and ImageNet, respectively. Methods that only adjust thresholds (e.g., LT-SNN) or only optimize SG (e.g., DeepTAGE) both perform worse than ours, indicating that the two challenges arising from membrane potential shifts limit performance. Enhancing SNNs requires considering both adaptive regulation of neuronal firing thresholds and dynamic optimization of gradient learning. In particular, as analyzed in ablation studies, most methods related to gradient optimization are superior to methods for threshold adaptation. Although AT-LIF$\&$ASG-S also achieves threshold and SG adjustment, it ignores the correlation between them, lacking synergistic learning. In summary, our method combines these two techniques to develop a two-stage synergistic learning that achieves higher accuracy with low latency, demonstrating weak-to-strong generalization and computational efficiency.

\subsection{Spike Firing Rates}
\begin{figure}[!ht]
	\centering
	\includegraphics[width=0.92\linewidth]{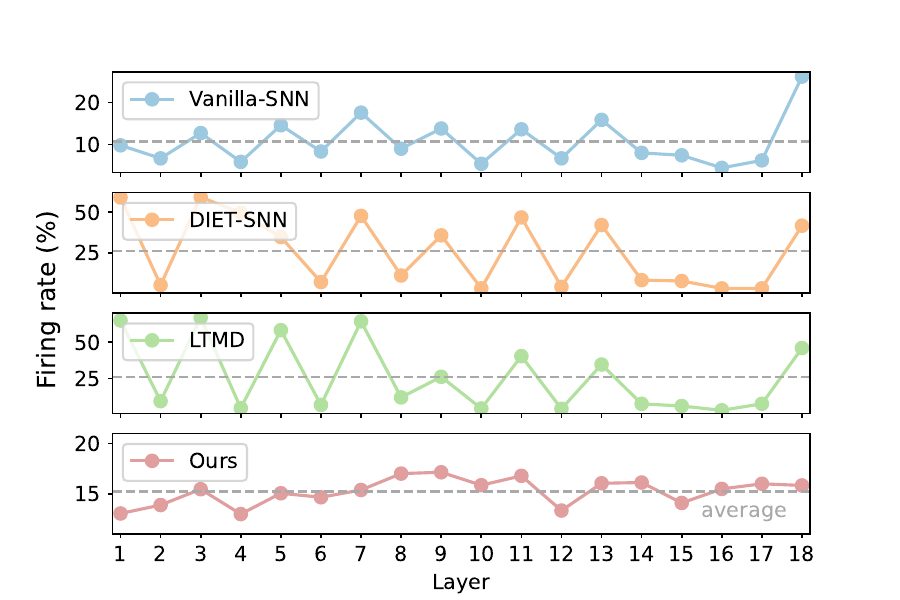}
	\caption{Comparison of firing rates for each layer in ResNet-19, where grey dotted lines denote average values.}
	\label{Fig::firing_rates_layers}
\end{figure}

To validate whether our method can effectively balance firing rates, we conducted experiments on the CIFAR10 dataset. In Fig.~\ref{Fig::firing_rates_layers}, Vanilla-SNN exhibits a low firing rate (average 10.66\%) with large fluctuations between layers, which hinders adequate encoding of input information. DIET-SNN \cite{rathi2023diet} and LTMD \cite{wang2022LTMD} methods substantially increase the firing level of neurons by adapting thresholds, avoiding extensive information loss. However, their firing rates across layers exhibit even greater oscillations, and the high firing rates certainly increase energy consumption.
\begin{figure*}[!ht]
	\centering
	\includegraphics[width=0.82\linewidth,height=3.8cm]{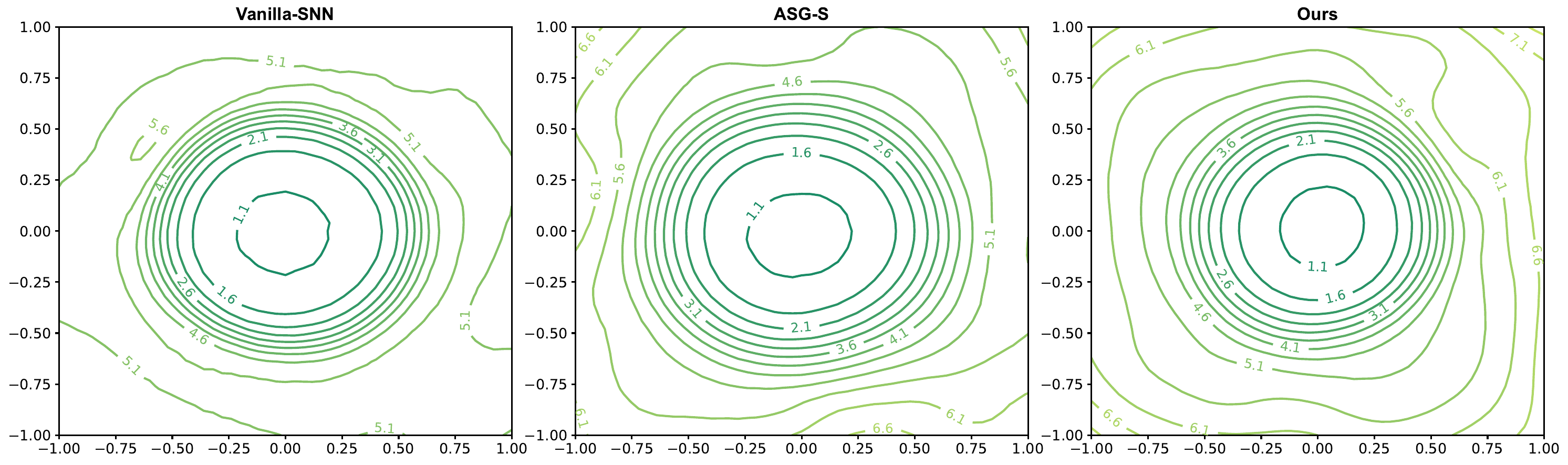}
	\caption{The 2D loss landscape of ResNet-19 trained with different methods on the CIFAR100 dataset.}
	\label{Fig::test_loss_2dcontour}
\end{figure*}
\begin{figure}[!ht]
	\centering
	\includegraphics[width=0.90\linewidth,height=6.1cm]{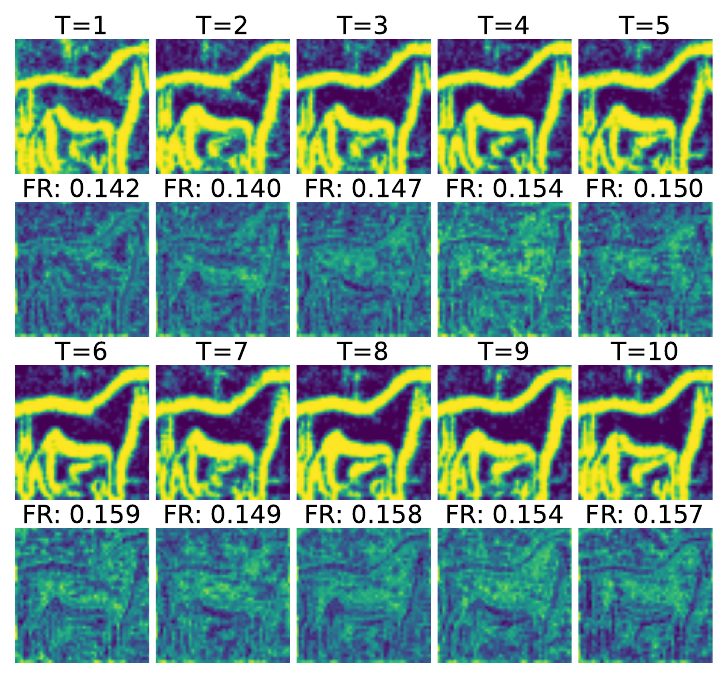}
	\caption{Case study of firing rates (FR) at each timestep.}
	\label{Fig::firing_rates_timesteps}
\end{figure}
Instead, our firing rates remain within a small fluctuation range of nearly 15$\pm$1.62\%. This stability stems from DS-ATGO's adaptive adjustment of thresholds based on the MPD distribution at each timestep, which helps neurons maintain moderate activation. Moreover, the firing rates of the last layer in the other three methods exhibit a steep increase. This may be due to the accumulated deviation of membrane potentials from thresholds over time, resulting in a slow decrease in firing rates. To compensate for the diminishing useful features in the hidden layers, output layer neurons are forced to fire more spikes by dynamically lowering thresholds or enhancing weights in an attempt to reconstruct complete feature representations. Although this temporarily enhances the excitability of output neurons, it causes neurons to over-respond to non-sharp signals, which is not conducive to recognition. Then, we presented the original image (rows 1, 3) and feature maps (rows 2, 4) on the CIFAR10-DVS dataset. In Fig.~\ref{Fig::firing_rates_timesteps}, our method effectively maintains the firing rate stable over timesteps and extracts good feature maps.

\subsection{Proportion of Gradient Available}
\begin{figure}[htbp]
	\centering
	\includegraphics[width=0.90\linewidth]{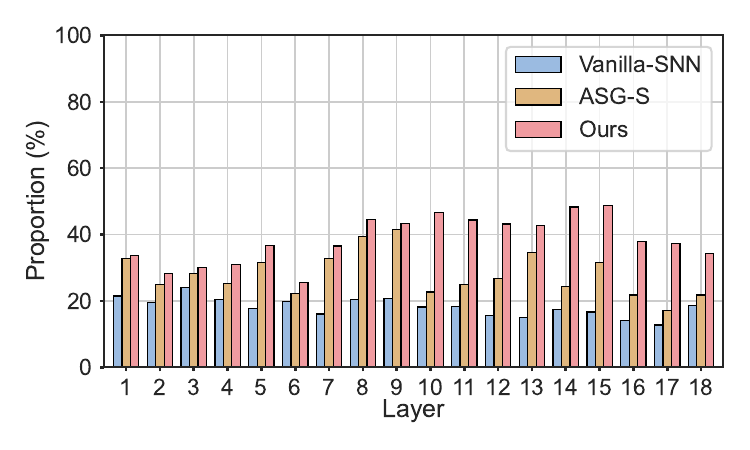}
	\caption{The proportion of neurons falling within the gradient-available interval of each layer in ResNet-19.}
	\label{Fig::gradient_available_rates}
\end{figure}

To investigate whether our method can effectively mitigate the gradient vanishing problem, we conducted experiments on the CIFAR100 dataset. In Fig.~\ref{Fig::gradient_available_rates}, the fixed SG of Vanilla-SNN cannot dynamically match the membrane potential shifts, resulting in only a few neurons in each layer falling within the gradient-available interval. As the number of layers deepens, this exacerbates the loss of gradient information, causing the proportion of neurons that obtain gradients in the last three layers to drop to an average of less than 15.13\%. Although ASG-S \cite{ai2025cross} slightly increases the proportion by adjusting SG, it still fails to accurately capture MPD that evolves with timesteps due to the lack of temporal dependence. By comparison, our method increases the gradient-available rate of each layer in ResNet-19 to over 38.53\%, and even maintains 36.54\% in the deeper layers.

Then, we visualized the 2D loss landscapes of these methods. In Fig.~\ref{Fig::test_loss_2dcontour}, the loss surface of Vanilla-SNN exhibits two local minima, with obvious protrusions in the contour lines around the center point, reflecting oscillations in weight updates caused by gradient loss. Although the loss landscape of ASG-S contains only one local minimum, its contour lines form an elliptical region. In contrast, the loss surface of our method exhibits a uniformly continuous downward trend in the parameter space through synergistic optimization, generating a sparser and flatter loss landscape.

\section{Conclusion}
In this paper, we present a new perspective on the performance limitations of SNNs by analyzing shifts in membrane potential distributions and deviations from the threshold. Fixed thresholds and SG are inherently mismatched with evolving MPD across timesteps, leading to imbalanced firing and diminished gradient problems. Then, we propose the DS-ATGO learning algorithm, which responds to evolving MPD across both temporal and spatial dimensions via forward adaptive thresholding and backward SG optimization. This dual-stage synergistic learning framework aims to help SNNs break through the performance bottlenecks of single-path techniques. Experimental results and analyzes demonstrate that DS-ATGO outperforms methods that only adjust thresholds or optimize SG. By balancing the firing rate and enhancing the gradient signal across timesteps, our method improves information encoding efficiency while optimizing loss landscapes, potentially advancing the application of SNNs to more complex tasks and wider scenarios.

\subsection*{Acknowledgements}
This work was supported by the National Key R$\&$D Program of China (Grant No. 2025YFG0100700) and the National Natural Science Foundation of China (Grant No. 62276235).

\bibliography{references}

\newpage
\maketitle
\appendix
\setcounter{secnumdepth}{1}
\setcounter{theorem}{0}
\setcounter{equation}{11}
\setcounter{table}{1}
\setcounter{figure}{11}

\twocolumn[{
	\begin{center}
		\Large\textbf{Appendix of \\ DS-ATGO: Dual-Stage Synergistic Learning via Forward Adaptive Threshold and \\ Backward Gradient Optimization for Spiking Neural Networks} \\[3em]
	\end{center}
}]

\section{Proofs of Theorems}\label{Sec::proofs_theorems}
\begin{theorem} 
	For $U(t)$ that satisfies a normal distribution $N(\mu,\sigma^2)$, the probability that a random variable $U_i(t)$ exceeds $\mu+\sigma$ is relatively constant and given by $P(U_i(t) > \mu+\sigma)=1-\varPhi(1)$, where $\varPhi(\cdot)$ denotes the cumulative distribution function of standard normal distributions.
\end{theorem}

\begin{proof}
	The probability density function of the normal distribution $U(t) \sim N(\mu,\sigma^2 )$ can be expressed by
	\begin{equation}
		f(U_i(t)) = \frac{1}{\sqrt{2\pi}\sigma} e^{-\frac{(U_i(t)-\mu)^2}{2\sigma^2}}, U_i(t) \in (-\infty,+\infty).
	\end{equation}
	To calculate the probability $P(U_i(t) > \mu+\sigma)$ that $U_i(t)$ is greater than $\mu+\sigma$, which is
	\begin{equation}
		P(U_i(t) > \mu+\sigma) = \int_{\mu+\sigma}^{+\infty} \frac{1}{\sqrt{2\pi}\sigma} e^{-\frac{(U_i(t)-\mu)^2}{2\sigma^2}}dU_i(t).
	\end{equation}
	
	Let $U(t)$ be normalized to $Z(t)=\frac{U(t)-\mu}{\sigma}$, we can obtain the standard normal distribution $Z(t) \sim N(0,1)$. Then, it transforms the calculation $P(U_i(t) > \mu+\sigma)$ into $P(Z_i(t) > 1)$, as follows:
	\begin{equation}
		P(Z_i(t) > 1) = \int_{1}^{+\infty} \frac{1}{\sqrt{2\pi}} e^{-\frac{Z_i(t)^2}{2}}dZ_i(t).
	\end{equation}
	Based on the properties of standard normal distributions, $P(Z_i(t) > 1) = 1 - P(Z_i(t) \le 1) = 1 - \Phi(1)$, where $\Phi(\cdot)$ denotes the cumulative distribution function. $\Phi(1)$ is a relatively constant value that does not depend on $\mu$ and $\sigma$.
\end{proof}

\section{Overall procedure of DS-ATGO algorithm}\label{Sec::overall_procedure}
\begin{algorithm}[!ht]
	\caption{The overall procedure of SNNs in one epoch with DS-ATGO algorithm.}
	\textbf{Input:} Initial SNN model and hyperparameters; total train and test iteration $N_{train}, N_{test}$; class label vector: $Y$. \\
	\textbf{Output:} The well-trained SNN and classification accuracy. \\
	\textbf{Training:}
	\begin{algorithmic}[1]
		\For{$n = 1$ to $N_{train}$}
		\For{$l = 1$ to $L-1$}
		\State Compute $I^l, U^l$ // (1) and (2)
		\For{$t = 1$ to $T$}
		\State Calculate $\Delta{V_{th}}^l(t)_n, \Delta{V_{th}}^l(t)$ // (5) and (6)
		\State Compute $S^l(t) \gets U^l(t) \geq \Delta{V_{th}}^l(t)_n$ // (3)
		\State Calculate the width of SG $k^l(t)$ // (10)
		\EndFor
		\EndFor
		\State Compute the output $O_n^L$ and loss $L_{CE}$ // (11)
		\State Compute the gradient w.r.t. $w$ // (13-15)
		\State Update $w \gets w - \eta \frac{\partial L}{\partial w}$, where $\eta$ is learning rate
		\EndFor
	\end{algorithmic}
	
	\textbf{Inference:}
	\begin{algorithmic}[1]
		\For{$n = 1$ to $N_{test}$}
		\For{$l = 1$ to $L-1$}
		\State Compute $I^l$ and $U^l$ // (1) and (2)
		\State Compute $S^l \gets U^l \geq \Delta{V_{th}}^l$ // (3)
		\EndFor
		\State Compute $O^L = \frac{1}{T} \sum^T_{t=1} (w^L \otimes S^{L-1})$ // (11)
		\State Compute accuracy from outputs $O^L$ and labels $Y$
		\EndFor
	\end{algorithmic}
\end{algorithm}

\section{Supplementary Experiments}
\subsection{Energy Efficiency and Ablation Study of Firing Control $f_c$}
We show the spike firing rates of different methods for each ResNet-19 layer in the \textit{Spiking Firing Rates} Section. Then, we estimate the energy consumption of ANN and various SNN methods ($T=2$). Following \cite{rathi2023diet}, the number of operations in SNN is specified as $FR^l \times T \times OP^l_{ANN}$, where $FR^l$ denotes the firing rate of the $l$-layer, $T$ denotes the timesteps and $OP^l_{ANN}$ denotes the number of operations at the $l$-th layer in the iso-architecture ANN. It is worth mentioning that, as the original images are directly fed into SNN for encoding and the membrane potentials of the output layer are used for prediction in our model, the number of operations for these two layers is specified as the number of operations in the corresponding layers of the iso-architecture ANN and scaled by $T$. \cite{horowitz20141} measured in 45-nm CMOS technology that an AC operation costs 0.9$pJ$ and a MAC operation costs 4.6$pJ$.

\begin{figure}[htbp]
	\centering
	\includegraphics[width=0.90\linewidth]{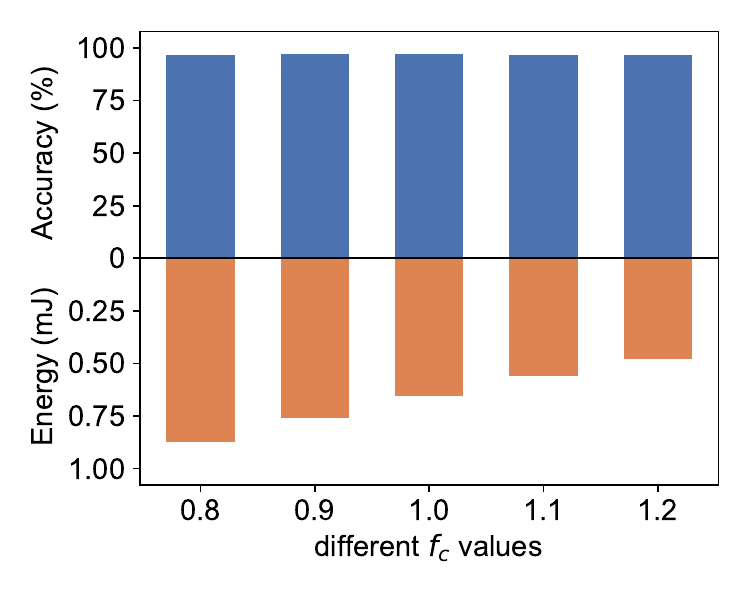}
	\caption{The trade-off between performance and energy consumption with different values of $f_c$.}
	\label{Fig::ablation_accuracy_energy}
\end{figure}

\begin{table}[!ht]
	\centering
	\setlength{\tabcolsep}{0.5mm}
	\renewcommand{\arraystretch}{0.8}
	\begin{tabularx}{\linewidth}{cccrcc}
		\toprule
		\textbf{Methods}
		& \textbf{AC}     & \textbf{MAC}       & \textbf{Energy}      & \textbf{Train}  & \textbf{Test}\\  
		\midrule
		ANN   & 0M & 2285.35M & 10.51mJ    & 38s  & 2s \\
		\midrule
		Vanilla-SNN & 443.14M & 7.08M & 0.43mJ    & 67s  & 4s\\
		\midrule
		DIET-SNN   & 1084.02M & 7.08M   & 1.01mJ   & 69s & 4s\\
		\midrule
		LTMD  & 1047.98M & 7.08M    & 0.98mJ    & 68s & 4s\\
        \midrule
		Ours ($f_c=1.0$)  & 690.84M & 7.08M    & 0.65mJ   & 74s & 4s\\
		\midrule
		Ours ($f_c=1.2$)  & 493.32M & 7.08M    & 0.48mJ   & 74s & 4s\\
		\bottomrule
	\end{tabularx}
	\caption{The energy consumption and running time of ANN and various SNN methods ($T=2$) on the CIFAR10 dataset.}
	\label{Tab::energy_consumption}  
\end{table}

To investigate the impact of the firing rate control factor $f_c$ on the trade-off between performance and energy consumption, we present a comparison of performance and energy consumption under different values of $f_c$, as shown in Fig.~\ref{Fig::ablation_accuracy_energy}. As listed in Table~\ref{Tab::energy_consumption}, the energy consumption of our method is approximately 0.65$mJ$ when the firing control factor $f_c=1.2$, much lower than the 10.51$mJ$ of ANN (around 16$\times$). When $f_c=1.2$, our energy consumption is comparable to Vanilla-SNN, while performance still improves by 1.06\%. Moreover, we recorded the running time of these methods. As we need to compute the thresholds and widths of SG for each timestep in each layer, the training time of our method is roughly 74s per epoch, which is 7 seconds slower than Vanilla-SNN. However, the inference time of our method is the same as that of the Vanilla-SNN, meaning that it does not introduce additional inference overhead.

\subsection{Ablation Study of Momentum Coefficient $m$}
\begin{figure}[!ht]
	\centering
	\includegraphics[width=1.0\linewidth]{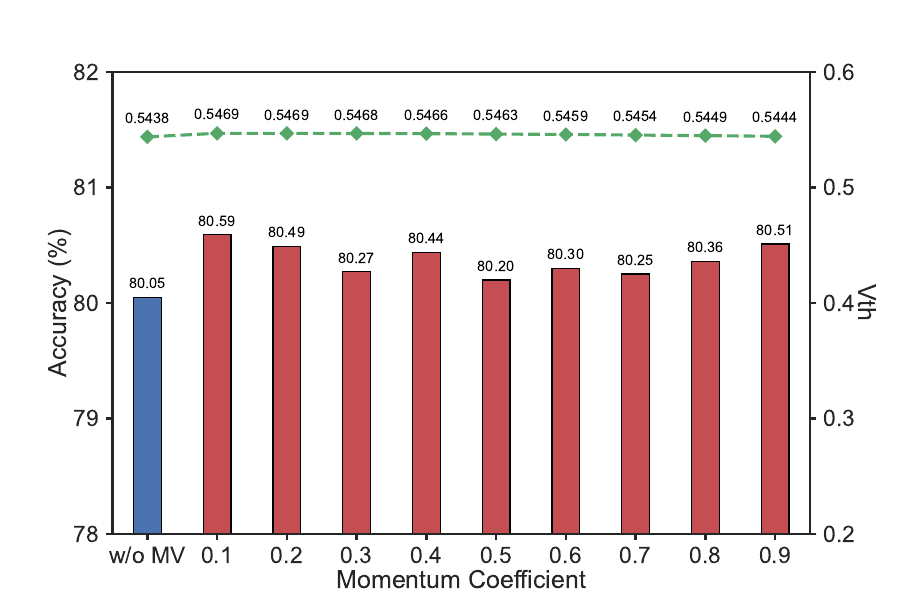}
	\caption{The impact of momentum coefficients in the moving average (MV) rule (bar chart) and a comparison of the moving average thresholds used for inference, obtained after training on the CIFAR100 dataset with different momentum coefficients (line chart).}
	\label{Fig::ablation_moving_average}
\end{figure}

In the \textit{Ablation Study} Section, we demonstrate that applying the moving average rule in the adaptive threshold mechanism can effectively improve the inference performance. Here, we further investigated the impact of the momentum coefficient on the moving average rule. As shown in Fig.~\ref{Fig::ablation_moving_average} (bar chart), different values of the momentum coefficient exhibit negligible influence on the accuracy of the inference phase, but all outperform the baseline without the moving average rule.
\begin{figure}[!ht]
	\centering
	\includegraphics[width=1.0\linewidth]{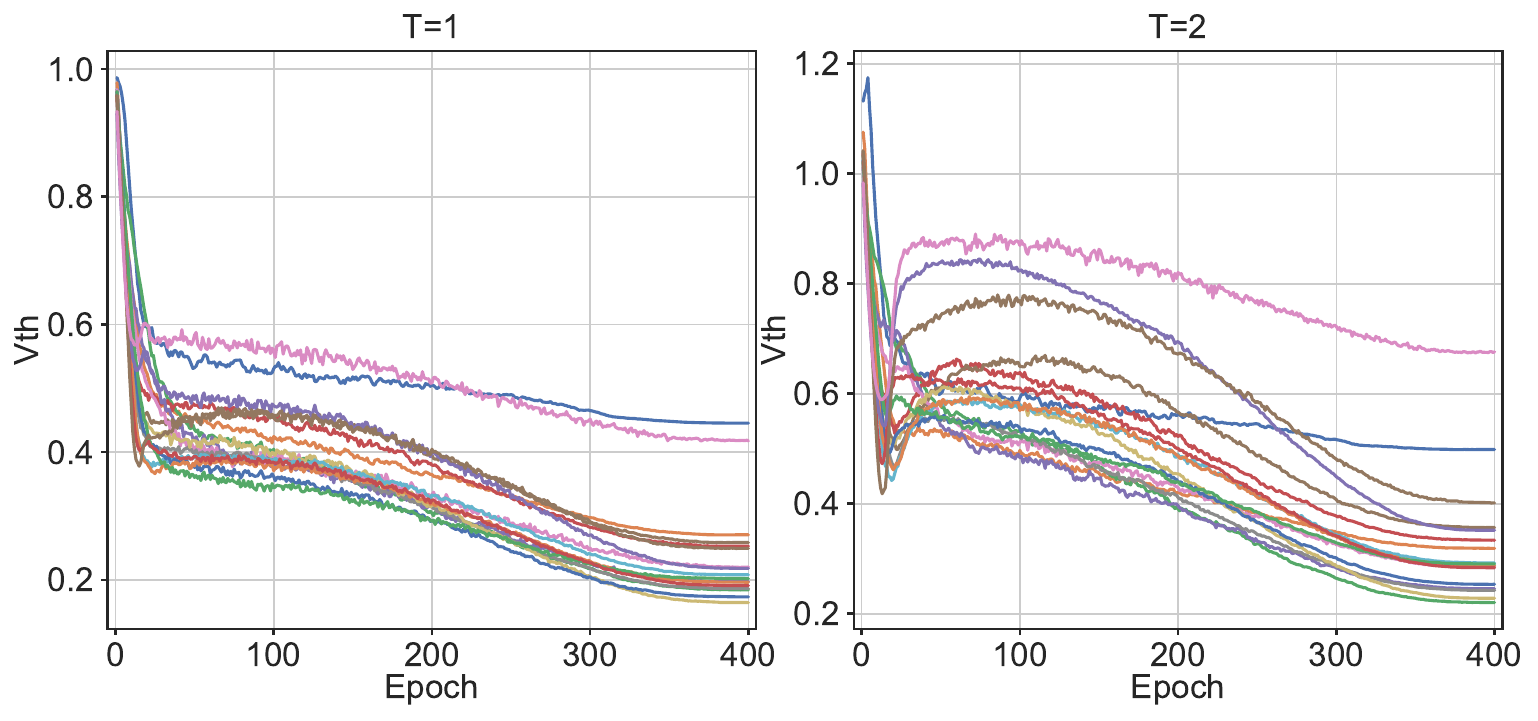}
	\caption{The changes in the thresholds (Eq.~\ref{Eq::adaptive_threshold}) of each layer in ResNet-19 during training on the CIFAR100 dataset.}
	\label{Fig::running_threshold}
\end{figure}
To explore the underlying reason for this phenomenon, we also present a comparison of the moving average thresholds (average of two timesteps) used for inference, obtained after training with different momentum coefficients. Fig.~\ref{Fig::ablation_moving_average} (green line chart) clearly shows that the moving average thresholds computed under different momentum factors exhibit a high degree of consistency. Finally, we plot the changes in the thresholds (Eq.~\ref{Eq::adaptive_threshold}) of each layer in ResNet-19 during training (Fig.~\ref{Fig::running_threshold}), where the momentum coefficient and $f_c$ are set to 0.1 and 1.0, respectively.

\subsection{Comparison of t-SNE}
\begin{figure}[ht]
	\centering
	\includegraphics[width=1.0\linewidth]{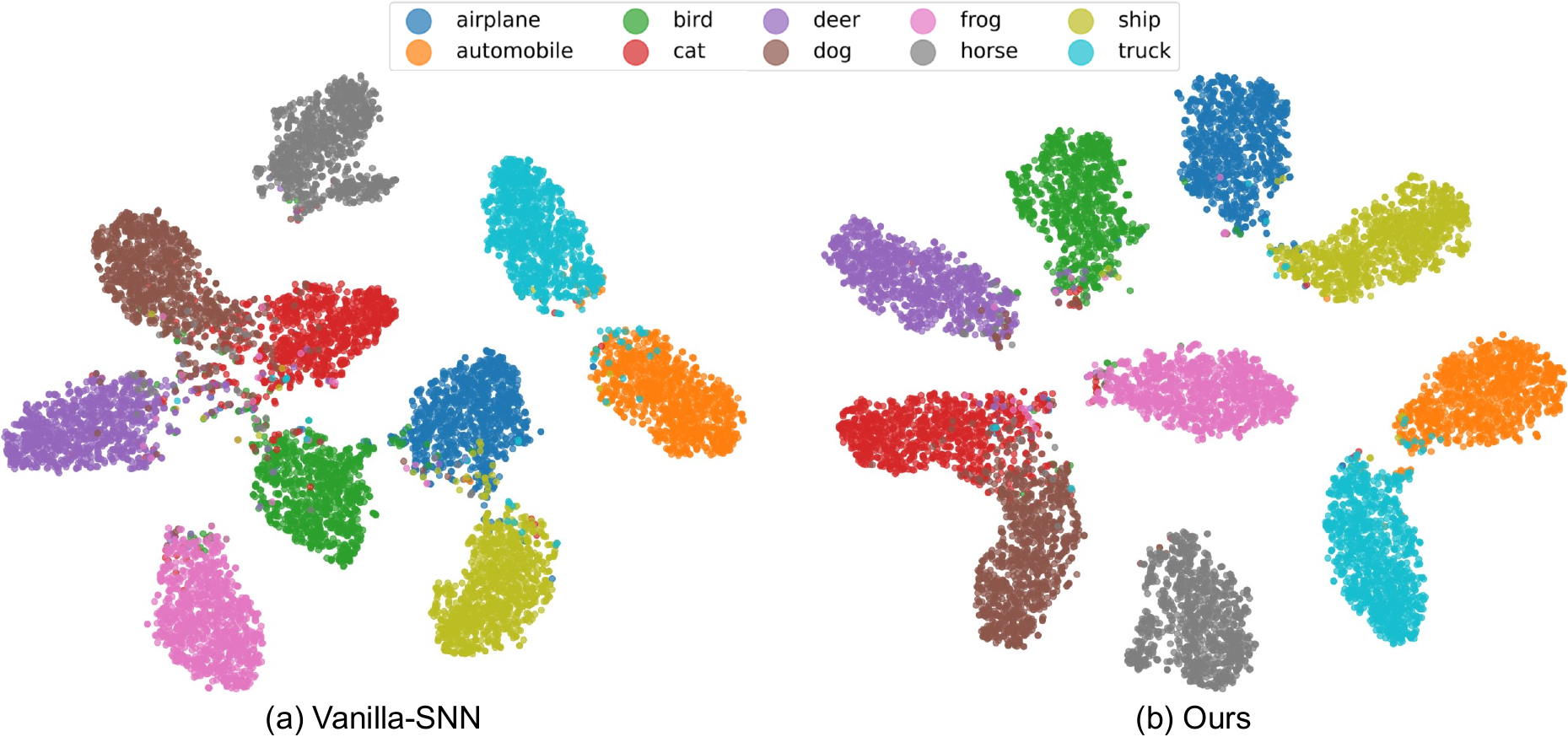}
	\caption{t-SNE visualization of feature vectors using ResNet-19 on the CIFAR10 dataset.}
	\label{Fig::t-SNE}
\end{figure}

In this section, we perform a t-SNE visualization on the feature vectors extracted from the last convolutional layer of ResNet-19. As shown in Fig.~\ref{Fig::t-SNE}, our method clearly distinguishes feature information of different categories. In particular, it can effectively separate similar categories such as ‘truck’ and ‘automobile’, while there is only partial overlap between the categories ‘dog’ and ‘cat’. However, Vanilla-SNN performs poorly in distinguishing features between multiple animal categories (such as ‘dog’, ‘cat’, ‘deer’, and ‘bird’). This indicates that our method has stronger feature extraction capabilities than Vanilla-SNN and effectively captures the crucial information of different categories, which is beneficial for classification.

\section{Implementation Details}
\subsection{Environment and Hyperparameter Settings}
All experiments are performed on a workstation equipped with Ubuntu 20.04.5 LTS, an AMD Ryzen Threadripper 3960X CPU running at 2.20 GHz, 128 GB RAM, and four NVIDIA GeForce RTX 4090 GPUs with 24GB DRAM. The code is implemented in the PyTorch framework with Python 3.9, and the weights are initialized using the default Kaiming initialization method in PyTorch 1.12.1.

\begin{table}[htbp]  
	\centering
	\resizebox{1.0\linewidth}{!}{
		\begin{tabular}{ccccc}
			\toprule
			Hyperparameters           &CIFAR10 &CIFAR100 &ImageNet &CIFAR10-DVS\\  
			\midrule
			$V_{th}$                 & 1.0           & 1.0       & 1.0    & 1.0    \\  
			$\tau$                   &  0.2          &  0.2       &  0.2    & 0.2    \\  
			Epoch                    &  400          & 400        &  400    &  400    \\  
			Batch Size                &  100           &  100       &   32   &  50    \\
			Optimizer                 &  SGD          &  SGD       &  SGD  &  SGD     \\  
			Weight Decay                 &  $1e^{-4}$          &  $1e^{-4}$       &  $1e^{-4}$  &  $1e^{-4}$     \\
			$\eta$             & 0.1           & 0.1        & 0.1 & 0.1      \\
			\bottomrule
		\end{tabular}
	}
	\caption{Hyperparameter Settings.}
	\label{Tab::hyperparmeters}
\end{table}
Table \ref{Tab::hyperparmeters} lists the hyperparameters used in our work. SGD optimizer with an initial learning rate $\eta = 0.1$, $0.9$ momentum, and weight decay $1e-4$ is used in all four datasets. All experiments used the CosineAnnealingLR scheduler to adjust $\eta$, which will cosine decay to 0 over epochs. The classification accuracy on the CIFAR10/100 and CIFAR10-DVS datasets are obtained by running with five random seeds, whereas that on ImageNet dataset is obtained using three random seeds.
For the CIFAR10/100 and ImageNet datasets, the network loss $L$ is calculated using
the cross-entropy function with label smoothing \cite{hu2024advancing}. For the CIFAR10-DVS dataset, the network loss $L$ is calculated using TET loss \cite{deng2022temporal}.

\subsection{Datasets and Preprocessing}
\textbf{CIFAR-10:} The CIFAR-10 dataset \cite{krizhevsky2009learning} consists of 60,000 RGB static images across 10 classes, each with a 32 × 32 pixels resolution. These images are split into 50,000 for training and 10,000 for testing. In data preprocessing, we normalized the dataset by subtracting the global mean value of pixel intensity and dividing by the standard variance of RGB channels. Each image was randomly cropped to 32 × 32 pixels after padding 4 pixels and randomly horizontally flipped with 0.5 probability. AutoAugment \cite{cubuk2019autoaugment} was used for data augmentation.

\textbf{CIFAR-100:} The CIFAR-100 dataset \cite{krizhevsky2009learning} also contains 60,000 RGB static images with a resolution of 32 × 32 pixels in 100 classes, which are split into 50,000 training images and 10,000 test images. We adopt the same preprocessing and data augmentation strategy to the CIFAR-100 dataset as the CIFAR-10 dataset.

\textbf{CIFAR10-DVS:} The CIFAR10-DVS dataset \cite{li2017cifar10} is converted from 10,000 CIFAR10 images and is the most challenging mainstream neuromorphic dataset. It consists of 10 classes, each with 1,000 samples and a resolution of 128 × 128 pixels, but was not split into training and test sets. Following previous studies \cite{zheng2021going,deng2022temporal,samadzadeh2023convolutional}, we also used 90\% of the samples in each class for training and the other 10\% for testing. In data preprocessing, we reduced the temporal resolution by segmenting the event stream into 10 temporal blocks, accumulating spike events within each block, and resizing the spatial resolution to 48 × 48 \cite{li2022neuromorphic,lian2023learnable}. Horizontal flipping with a random probability greater than 0.5 and rolling with 5 offsets were applied for data augmentation \cite{deng2022temporal}.

\textbf{ImageNet:} The ImageNet-1k dataset \cite{deng2009imagenet} is more challenging than static CIFAR family datasets. It consists of 1250k RGB static images for training and 50k RGB static images for testing across 1000 classes. In data preprocessing, we normalized the dataset by subtracting the global mean value of pixel intensity and dividing by the standard variance of RGB channels. Each image was randomly cropped to 224 × 224 pixels and randomly horizontally flipped with 0.5 probability. AutoAugment was used for data augmentation.

\subsection{Network Architectures}
We adopt the widely-used ResNet-18/19 \cite{zheng2021going,fang2021deep,hu2024advancing} and VGGSNN \cite{ deng2022temporal} network structures.The details of the network architecture are listed in Table \ref{Tab::ResNet-19} and Table \ref{Tab::VGGSNN}, respectively. $xCy$ represents a convolutional layer with output channels equal to $x $, $kernel~size = y$, $stride~set = 1$, and $padding = 1$. $xFC$ represents a fully-connected layer with output features equal to $x$. $2AP$ represents the average-pooling layer with $kernel~size = 2$ and $stride = 2$. $z$ is the number of classes.

\begin{table}[htbp]
	\centering
	\begin{tabular}{cc}
		\toprule
		conv1 & 128C3 \\
		\midrule
		block1 & $\left( \begin{array}{c}128C3\\128C3\end{array}\right)\times3$ \\
		\midrule
		block2&$\left( \begin{array}{c}256C3\\256C3\end{array}\right)^*\times3$ \\
		\midrule
		block3&$\left( \begin{array}{c}512C3\\512C3\end{array}\right)^*\times3$ \\
		\midrule
		ResNet-18& average pool, $z$FC \\
		\midrule
		\multirow{2}*{ResNet-19}& average pool, 256-d FC \\
		&  $z$FC \\
		\bottomrule
	\end{tabular}
	\begin{tablenotes}
		\footnotesize
		\item * means the first basic block in the series performs downsampling directly with convolution kernels and a stride of 2.
	\end{tablenotes}
	\caption{ResNet-18/19 structures.}
	\label{Tab::ResNet-19}
\end{table}

\begin{table}[htbp]
	\centering  
	\resizebox{0.75\linewidth}{!}{
		\begin{tabular}{cc}
			\toprule
			\multirow{6}{*}{VGGSNN} & 64C3-LIF \\
			& 128C3-LIF-2AP \\ 
			& 256C3-LIF-256C3-LIF-2AP \\
			& 512C3-LIF-512C3-LIF-2AP \\
			& 512C3-LIF-512C3-LIF-2AP \\
			& -$z$FC \\
			\bottomrule
		\end{tabular}
	}
	\caption{VGGSNN structures.}
	\label{Tab::VGGSNN}
\end{table}

\end{document}